\documentclass{CVM}

\def\sysName{\textit{PartHOI}}

\CVMsetup{
type      = {Research Article},
doi       = {s41095-0xx-xxxx-x},
title     = {\sysName: Part-based Hand-Object Interaction Transfer via Generalized Cylinders},
author    = {Qiaochu Wang$^{1}$, Chufeng Xiao$^{1}$, Manfred Lau$^{2}$, and Hongbo Fu\cor{}$^{3}$\\},
runauthor = {Q. Wang, C. Xiao, M. Lau , H. Fu},
abstract  = {
{

Learning-based methods to understand and model hand-object interactions (HOI) require a large amount of high-quality HOI data. One way to create HOI data is to transfer hand poses from a source object to another based on the objects' geometry. However, current methods for transferring hand poses between objects rely on shape matching, limiting the ability to transfer poses across different categories due to differences in their shapes and sizes. We observe that HOI often involves specific semantic parts of objects, which often have more consistent shapes across categories. In addition, constructing size-invariant correspondences between these parts is important for cross-category transfer. Based on these insights, we introduce a novel method \sysName~ for part-based HOI transfer. Using a generalized cylinder representation to parameterize an object parts’ geometry, \sysName~ establishes a robust geometric correspondence between object parts, and enables the transfer of contact points. Given the transferred points, we optimize a hand pose to fit the target object well. Qualitative and quantitative results demonstrate that our method can generalize HOI transfers well even for cross-category objects, and produce high-fidelity results that are superior to the existing methods.}
},
keywords  = {Geometry analysis, object modeling, hand pose synthesis, hand-object interaction},
copyright = {The Author(s)},
}

\begin{document}

\maketitle

\begin{figure}[b] 
    \small\renewcommand\arraystretch{1.3}
        \begin{tabular}{p{80.5mm}} \toprule\\ \end{tabular}
        \vskip -4.5mm \noindent \setlength{\tabcolsep}{1pt}
        \begin{tabular}{p{3.5mm}p{80mm}}
    $1\quad $ & School of Creative Media, City University of Hong Kong, Hong Kong, China. E-mail: Q. Wang, qiaowang-c@my.cityu.edu.hk; C. Xiao, chufeng.xiao@my.cityu.edu.hk. \\
    $2\quad $ & Department of Computer Science, Lakehead University, Orillia, Ontario, Canada. E-mail: M. Lau, manfred.lau@lakeheadu.ca. \\
    $3\quad $ & Division of Arts and Machine Creativity, Hong Kong University of Science and Technology, Hong Kong, China. E-mail: H. Fu, hongbofu@ust.hk.\\
 \vspace{-2mm}
    \end{tabular} \vspace {-3mm}
    \end{figure}

\begin{figure*}[h!t]
  \centering
  \includegraphics[width=\textwidth, alt={PartHOI compare to other methods}]{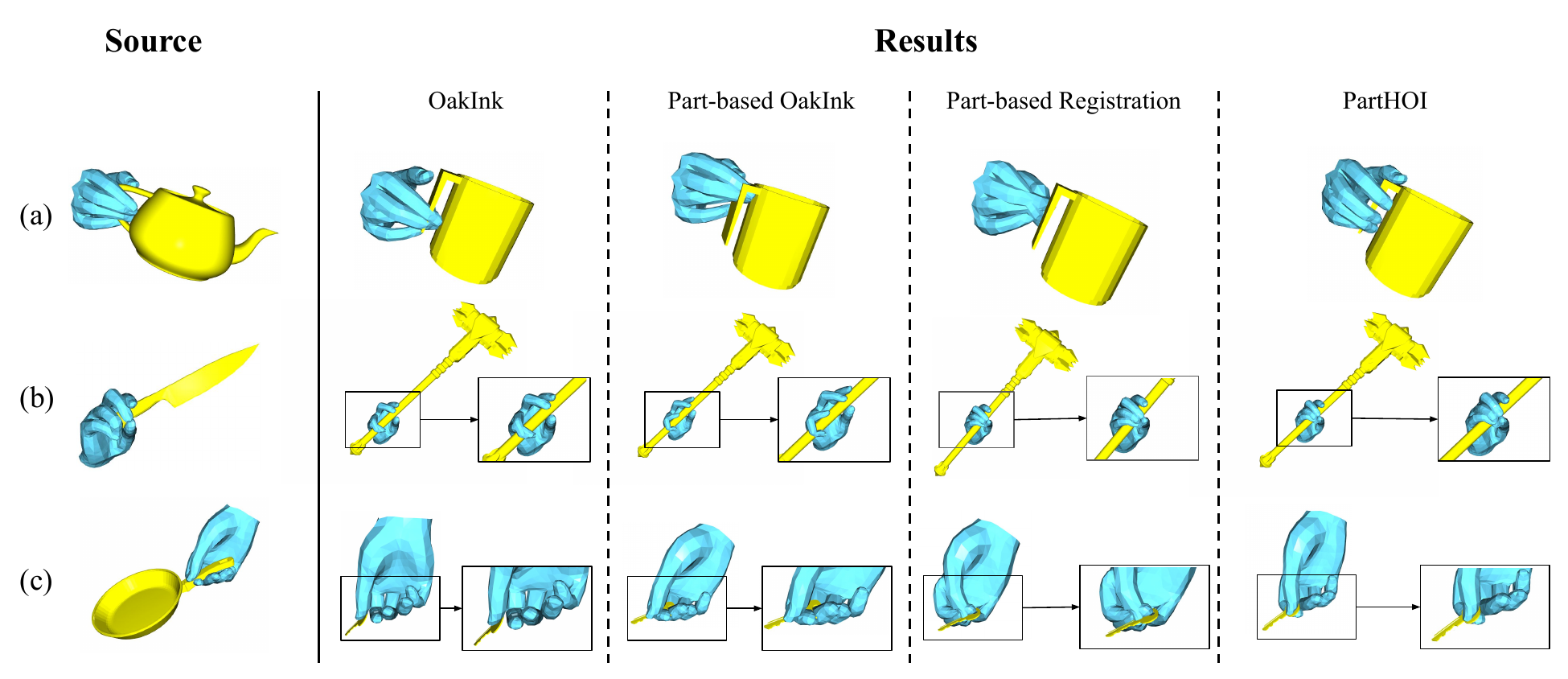}
  \caption{
  	We introduce \sysName, a pipeline for hand pose transfer, based on object parts represented by generalized cylinders. We transfer a hand pose from an existing hand-object interaction to a new object. Our work can transfer plausible hand grasps between cross-category objects, even with a large geometry difference (a, b) or a large size difference (c). In contrast, the existing work OakInk, its part-based version and, part-based direct transfer via correspondence would easily lose hand contact and produce unnatural hand poses if there are significant differences between the two objects. 
  }
  \label{fig:teaser}
\end{figure*}

\section{Introduction}
\label{sec:intro}
Understanding and simulating the human-object interaction (HOI) is a fundamental task for reconstructing realistic 3D worlds. Many learning-based methods have been proposed to address problems related to this topic, including 3D reconstruction \cite{cao_reconstructing_2021,hasson_towards_2021,hasson_learning_2019,tzionas_3d_2015,chen_gsdf:_2023,li_chord:_2023}, object localization \cite{chalon_online_2013,Lee_2020_WACV}, 3D object/hand tracking \cite{Tsoli_2018_ECCV,hamer_tracking_2009,leibe_real-time_2016,krainin_manipulator_2011,rangesh_hidden_2016,mueller_real-time_2017,panteleris_back_2017}, and grasp generation \cite{jiang_hand-object_2021,Brahmbhatt_2020_ECCV,GRAB:2020,brahmbhatt_contactgrasp:_2019,grady_contactopt:_2021,zhu_toward_2021,xu_unidexgrasp:_2023,liu_contactgen:_2023,brahmbhatt_contactdb:_2019,christen2022d}. However, such methods need a large-scale and high-quality HOI dataset for training. The collection of such a dataset demands high-precision motion capture (MoCap) equipment and significant labor resources.

\begin{figure*}[h!]
  \centering
  \includegraphics[width=\textwidth, alt={Fail case for baseline}]{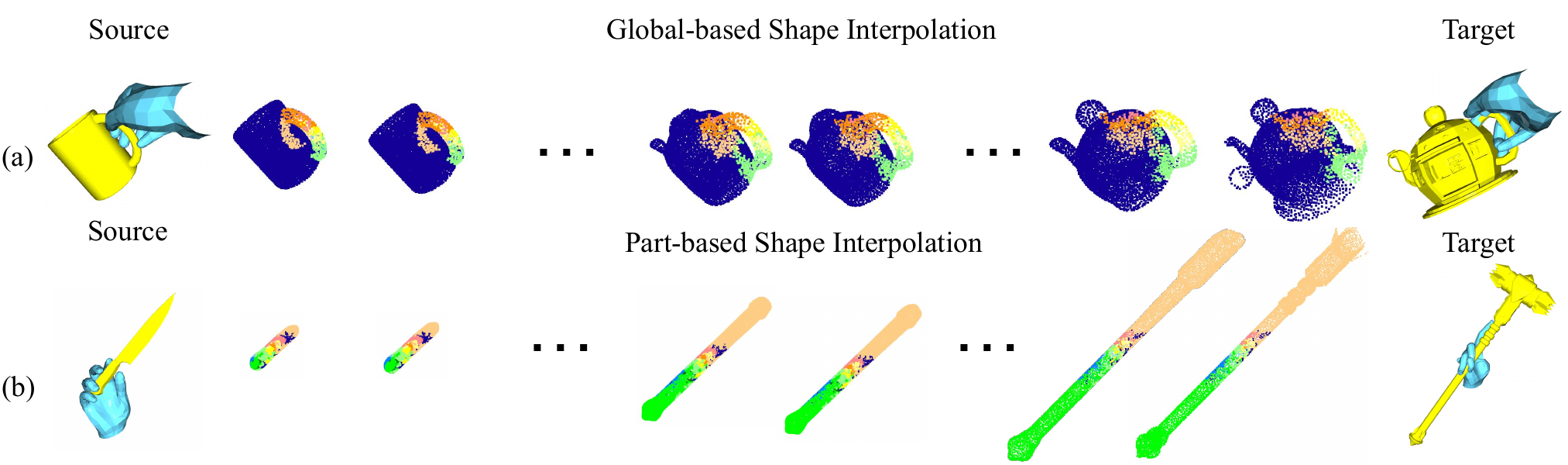}
  \caption{Examples of contact map transfers based on 
  (a) global-based shape interpolation and (b) part-based shape interpolation.}
  \label{fig:contact}
\end{figure*}

One possible way to acquire HOI data at a low cost is to produce a hand pose given the geometry of a target object.
Another method (OakInk \cite{yang_oakink:_2022})
transfers a hand pose from existing HOI data
(e.g., a hand pose grasping a source object scanned by MoCap) to a new object given the correspondence between the affordance maps (which indicates the contact points between a hand pose and an object) of the two objects. However, the method could not produce an accurate and plausible hand pose when the source object has a significant geometry difference from the target object, as shown in Figure \ref{fig:contact}a. This is because the transfer method highly relies on the whole shape of the two objects to compute the correspondences between them.
Thus, the method struggles to transfer a hand pose between cross-category objects (e.g., mug and teapot), even though they have similar affordance in a local part.

We observe that most object grasps only involve local object parts. Compared to a whole shape, a local part is a more generic medium and can provide greater generalization ability to transfer hand poses between cross-category objects. For example, a mug and a teapot may have a similar affordance part in their handles. However, applying shape-based transfer for local parts or directly extending OakInk to a part-based method still could not address the problem well. This is because the cross-category object parts with similar affordance may vary largely in size while shape-based transfer or OakInk is sensitive to object sizes since they build the correspondence between two shapes via shape registration or interpolation.
Moreover, parts from cross-category objects are more likely to have significant shape differences, further affecting the correspondences between object parts. Figure \ref{fig:contact}b shows a failure case of part-based OakInk that estimates inaccurate contact points and thus produces unreasonable hand poses.
Specifically, the local changes between the two handle parts of a knife and hammer, i.e., the length of the skeletons and the scale of the cross-sections, would significantly worsen the prediction of the hand poses. Figure \ref{fig:contact}a and \ref{fig:contact}b illustrate the failure cases of OakInk. Despite the object parts having similar affordances, significant shape differences lead to incorrect correspondences between the source and target.

Based on the above observations, we introduce a novel part-based transfer method via the generalized cylinder (GC) representation to solve the problem of cross-category HOI transfer.
Compared to the global-based transfer method, the part-based method can avoid the influence of shape differences in object regions that are irrelevant to the hand pose transfer. Additionally, using parts as the transferring units is more generic and allows for the use of parametric representations to represent contact regions. The generalized cylinder serves as a primitive representation of 3D object geometry, enabling the establishment of a unified and size-invariant correspondence between two different object parts.
Specifically, we first model a generalized cylinder to abstract and parameterize the object parts of a source and target. 
Similar to the representation employed in GCD \cite{gcd15}, a GC represents each point on the surface via the medial skeleton of an object part and cross-section profile curves along the skeleton. Following the polar coordinate systems created by the modeled GC, the contact points between a hand pose and a source part can be easily transferred to a target part. To maintain the target hand-grasping pose to be similar to the source pose, we estimate the target contact map while preserving the relative positions among contact points unchanged. Unlike methods that use registration or shape interpolation to find correspondences, the GC-based method mitigates the influence of size changes by preserving the relative positions among contact points. It also reduces the impact of shape differences by ensuring that contact points maintain their relative positions on the object globally. Finally, based on the re-targeted affordances, we refine a computed hand pose to approximately grasp the target object part through an optimization process, similar to the pose refinement method adopted in OakInk.

To the best of our knowledge, our method is the first work to adopt generalized cylinders for part-based HOI transfer, thus enabling generic cross-category transfer. To evaluate the performance of our method in cross-category HOI transfer, we prepare a new HOI dataset with a total of 120 source-to-target pairs, including source HOI data (objects and grasping poses) and target objects. The source data covers 16 categories of objects, while the target objects include 25 categories. In addition to the categories from OakInk, our dataset includes 8 additional object categories (such as can openers and toys) which, to our knowledge, are not in any existing hand-object interaction datasets. With the dataset, we tested our method and three baselines, i.e., vanilla OakInk, direct part-based transfer, and part-based OakInk (adapted for part-based transfer), respectively for intra-category transfer and cross-category transfer.
Through quantitative and qualitative comparisons, we demonstrate that \sysName~can produce more plausible and faithful hand poses and has higher generalization ability, especially for cross-category transfer.

\section{Related work}
\subsection{{3D Object Decomposition}}

3D object decomposition decomposes complex objects into simple and semantic parts. It plays a pivotal role in the understanding and manipulation of 3D structures.
For example, to train a learning-based method for hierarchical semantic segmentation, PartNet \cite{mo2019partnet} presents a large-scale 3D object benchmark with fine-grained part-level annotations. Unlike learning-based methods, GCD proposes an optimization method to decompose a 3D object into cylindrical parts following the principle of geometry simplicity. However, the decomposition of GCD only considers the cylindricity measurement of object geometry but not the affordance involved by human hands. Different from the object decomposition task that decomposes a single object into parts, our work focuses on constructing the affordance correspondence between two object parts. To ensure the correspondence is size-invariant and unified, we utilize a generalized cylinder representation (similar to GCD) to represent an object part.

\subsection{Object Affordance Estimation}
Object affordance segments daily objects into different regions based on how humans grasp the objects. AffordanceNet and 3D affordanceNet \cite{do_affordancenet:_2018,deng_3d_2021} propose learning-based methods trained on their prepared datasets to estimate object affordance given object shapes. Tactile Mesh Saliency \cite{Lau_tactile} generates grasp saliency maps based on the geometry and visual features of 3D objects.
However, such learning-based methods struggle to produce plausible affordance maps for out-of-domain data. To generate robust contact maps for optimizing a hand pose, we transfer a contact map from a source shape to a target shape, based on a unified representation of generalized cylinders.

\subsection{Hand Grasp Synthesis}
Many early works investigate grasp synthesis for robotics. Researchers use physics-based or deep learning methods to estimate grasping poses for robotic arms or grippers to grasp objects. GraspIt! \cite{miller_graspit!_2004} is an early open-source toolbox to generate grasps based on physical simulations of objects and grippers. SynGrasp \cite{6630708} provides faster simulation in MATLAB Simulink. Some works \cite{dang_semantic_2012,detry_task-oriented_2017} use semantic recognition in grasping scenes to produce semantic-aware robot poses. ObMan \cite{hasson_learning_2019} incorporates MANO \cite{MANO:SIGGRAPHASIA:2017}, a parametric hand model, with GraspIt! to generate grasps based on the collected dataset of human hand images. Although such robotic-oriented methods can be applied to human hand grasp synthesis by mapping poses from grippers to human fingers, the methods would often produce unnatural poses for human grasping due to the gap between humans and robots.

With the advancement of deep learning and reinforcement learning techniques recently, researchers aim to develop a data-driven model to generate grasps based on the geometry of 3D objects. For example, ObMan trains a deep neural network on their proposed dataset to jointly reconstruct a hand pose and a 3D object model given a monocular RGB image. Many works propose diverse representations of HOIs and adopt cVAE-like models \cite{kingma2013auto,rezende2014stochastic} to generate such representations based on object geometry features. Specifically, GrabNet \cite{GRAB:2020} and GraspTTA \cite{jiang_hand-object_2021} trains cVAE networks with well-captured datasets to sample hand poses or contact maps given the geometry features of 3D shapes. Contact2Grasp \cite{ijcai2023p117} presents a novel hand pose refinement method that divides a hand into multiple regions and separately optimizes the hand regions to approximate a given contact map. Yang et al. \cite{yang_cpf:_2021} use anchors to represent the local regions of a hand and project the closest anchors from the hand to the object surface to form a contact map. Contactgen \cite{liu_contactgen:_2023} introduces contact direction between a hand and an object surface to represent HOI data. Instead of generating static hand poses, ManipNet \cite{zhang_manipnet:_2021} generates a sequence of hand motions of interactions with objects based on hand wrist and object movements. Although these data-driven models can generate plausible results, they require high-quality data as input and lack controllability for hand pose outputs. In contrast, our method can produce a hand pose similar to a given pose to grasp a new object.

Reinforcement learning has been widely applied to grasp generation with the help of physical simulation engines. For example, She et al. \cite{she_learning_2022} generate high degrees of freedom (DOFs) grasping poses between robot grippers and objects via reinforcement learning, while D-Grasp \cite{christen2022d} synthesizes human hand motions. Although these data-driven methods can produce impressive hand poses, they highly rely on the training data and could not perform well for out-of-domain objects.

To solve the problem of HOI transfer for diverse categories of objects, we propose a generic grasp synthesis method to transfer a hand pose between two objects. Our method is optimization-based and eliminates the need for training data, thus mitigating the domain-gap issues existing in the learning-based methods.

\subsection{HOI data transferring}
Transferring HOI data from source to target is a solution that can take advantage of high-fidelity HOI data from an existing small-scale dataset. OakInk transfers a contact map \cite{yang_cpf:_2021} from a source object to a target object through shape interpolation by DeepSDF \cite{park_deepsdf:_2019}. However, as discussed in Section 1, this method suffers greatly from the issues caused by size and shape variance between source and target objects. TOCH \cite{zhou2022toch} can transfer interaction sequences between objects with similar geometry features by learning to refine HOI sequences via GRU modules \cite{chung2014empirical}. Similarly, we propose a transfer method to produce high-quality hand poses for grasping general objects. We compare our method with the existing transfer baseline (OakInk) to show its effectiveness and superiority (Section \ref{sec:exp}).

\section{Method}
\begin{figure*}[t]
  \centering 
  \includegraphics[width=\textwidth, alt={overview}]{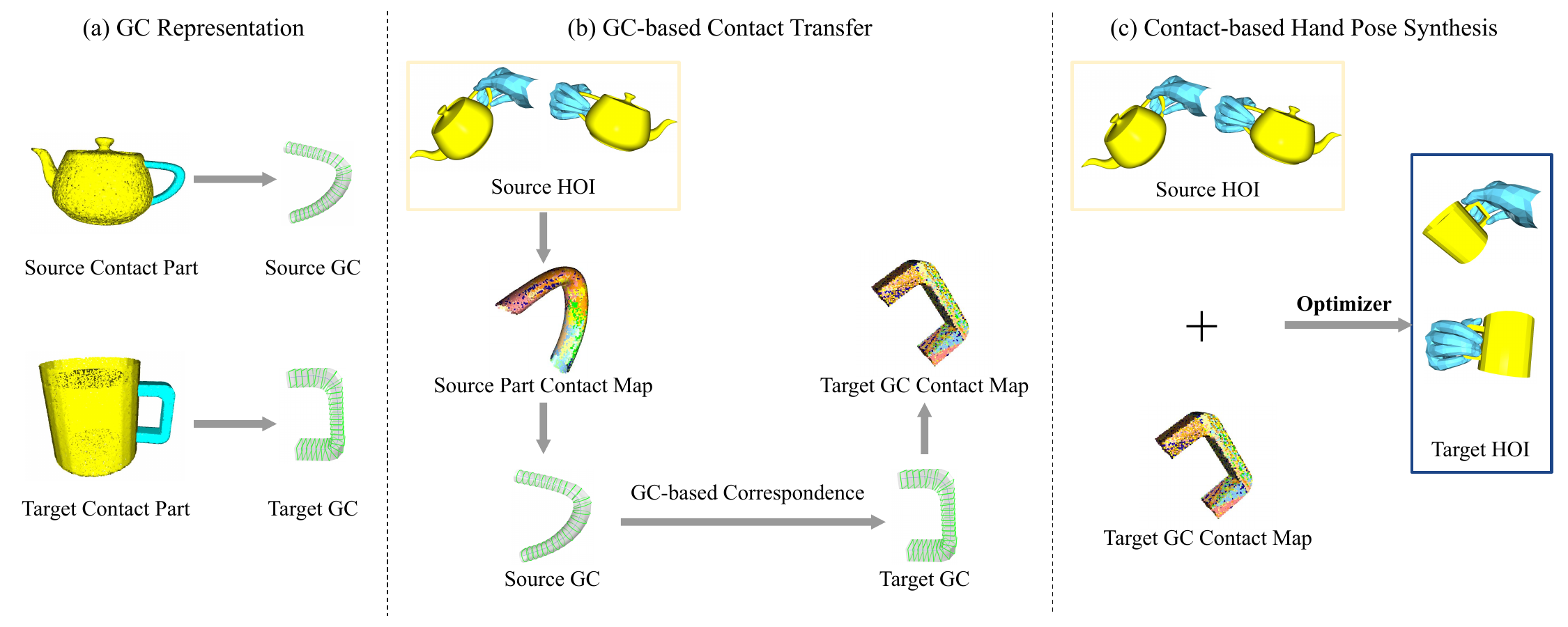}
  \caption{
  The pipeline of our \sysName~method. (a) We first model a given source and target object part using a generalized cylinder representation. (b) Next, we transfer the contact map from the source part to the target one via GC-based correspondence. (c) Following the target contact map, we adopt a hand optimizer similar to OakInk to optimize a hand pose initialized from the source hand pose.
  }
  \label{fig:overview}
\end{figure*}

Our goal is to produce high-quality hand poses by transferring the knowledge from existing HOI data to grasp a new object. Recognizing that a semantic local part of an object is a more generic unit than the whole object, we propose a part-based transfer method based on the generalized cylinder representation.
Figure \ref{fig:overview} shows an overview of our \sysName~pipeline. We first introduce generalized cylinders (GCs) to abstract the geometry of two object parts and form unified polar coordinate systems as the shape representation (Section \ref{sec:GC}). Next, we transfer a contact map from the source GC to the target GC following two formulated constraints (Section \ref{sec:contact_transfer}). Given the re-targeted contact map, we adopt an optimization method to refine the hand pose to grasp the target object (Section \ref{sec:pose_refine}). We describe the details of our method in the following subsections.

\subsection{Generalized Cylinder Representation}
\label{sec:GC}

\begin{figure}[h]
  \centering 
  \includegraphics[width=\columnwidth, alt={GC}]{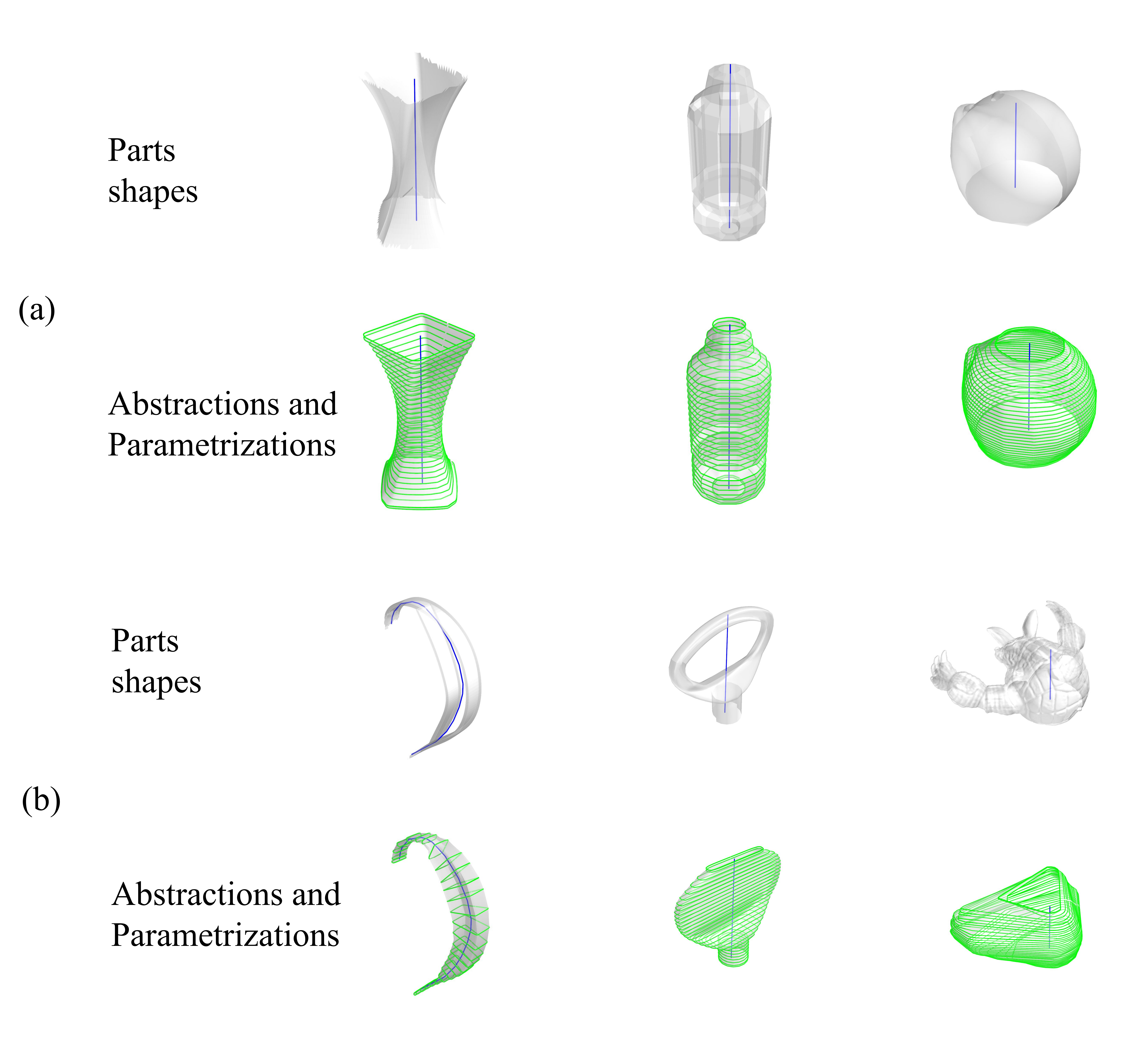}
  \caption{Abstract and parameterize object parts represented as GCs.
  }
\label{fig:gc_abstraction}
\end{figure}

\begin{figure}[h]
  \includegraphics[width=\columnwidth, alt={GC}]{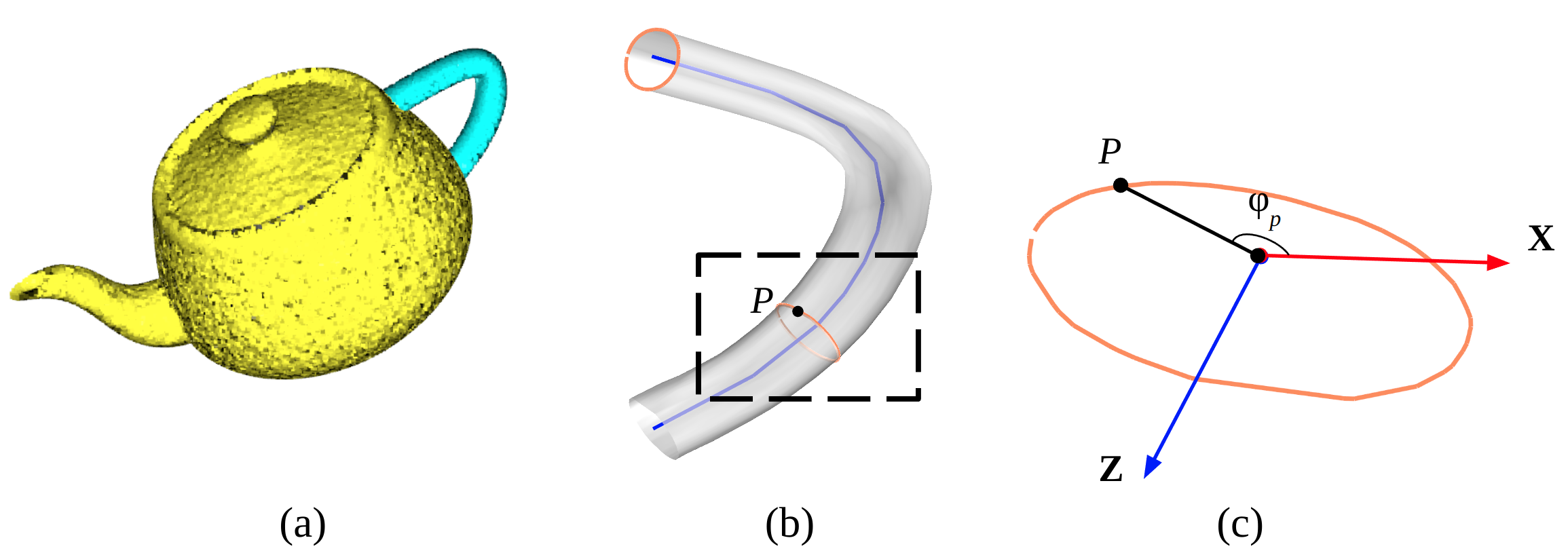}
  \caption{Given a segmented object part (highlighted in cyan in (a)), our method first abstracts and parameterizes the shape into a GC (b), which represents each surface point $p$ via a polar coordinate system (c).
  }
  \label{fig:gc_representation}
\end{figure}

Similar to GCD, we represent an object part using a generalized cylinder (GC) through the medial-axis skeleton (the blue line in Figure \ref{fig:gc_abstraction}) of the part and cross-section profile curves (the orange lines in Figure \ref{fig:gc_representation}b) along the skeleton. 
The object part and skeleton are manually segmented and extracted from a whole object following users' grasping preference.
Automated segmentation methods can also be used if desired.
Each cross-section curve can be automatically computed by determining the intersection points between a skeleton-along profile plane and the surface of the object part.

When we use our hands to grasp a specific, semantically meaningful part of an object, our grasp pose tends to align with the primary structure or overall shape of that part. Minor structural details or small features have less influence on the grasp, as the hand primarily adapts to the core shape needed for a stable hold. As shown in Figure \ref{fig:gc_abstraction}, some
object parts are directly representable by generalized cylinders (Figure \ref{fig:gc_abstraction}a), while other parts cannot be directly represented by generalized cylinders due to their complexity (Figure \ref{fig:gc_abstraction}b). In our work, geometric abstraction is applied to these object parts for approximation by generalized cylinders. This approach approximates the object's form by reducing complex features, such as intricate details, irregularities, or fine-scale variations, into simpler geometric structures. Instead of capturing all geometric details, we approximate the shape by filling holes, smoothing contours, or simplifying features, making the representation computationally efficient while still maintaining the essential structure of the part for hand grasping. With this geometric abstraction,
the resulting transferred contact map generally enables reasonable grasp poses. The GC parameters are based on the GC representation of object parts. Specifically, we divide the skeleton of an object part into $n$+1 regions by evenly sampling $n$ cross-section curves to form a GC. For each cross-section curve, we {represent it as a series of connected line segments that form a closed loop.} Empirically, we found that $n$=30 is enough to maintain the necessary geometry of an object for estimating a target hand grasp. A larger $n$ leads to limited improvement but much lower computation efficiency.

Given a single skeleton and $n$ cross-section curves, we establish a polar coordinate system $p(h,\varphi)$ to represent a surface point $p$ of an object part. Specifically, $h\in[0,H]$ indicates point height along the shape skeleton. The height $h$ is determined by projecting a surface point $p$ onto the object skeleton from the opposite direction of the surface normal.
As illustrated in Figure \ref{fig:gc_representation}c, $\varphi$ represents a relative angle between $\overrightarrow{op}$ and the x-axis on $h$-th cross-section plane, where $o$ is an intersection point between the object skeleton and the cross-section plane, while the x-axis is perpendicular to the medial longitudinal section. Based on the representation, we model a source GC and a target GC, respectively, from the source and target objects.

\subsection{GC-based Contact Transfer}
\label{sec:contact_transfer}

A contact map consisting of several contact points between hand joints and object surface can effectively reflect how a hand grasps an object in the HOI data. Thus, we set up a contact map \cite{yang_cpf:_2021} as a medium to bridge the correspondence between two partial shapes. Unlike OakInk-based methods that are sensitive to object shapes and sizes due to shape-interpolated correspondences (Figure \ref{fig:teaser}), we construct size-invariant correspondences via a GC representation to transfer a contact map between two GCs. We mainly describe below how to transfer the contact map from the source GC to the target GC. First, we transfer the contact map from a source shape to the modeled source GC by finding the closest point on the GC surface for each contact point. 

\begin{figure}[htb]
  \centering 
  \includegraphics[width=.8\linewidth, alt={surface_point}]{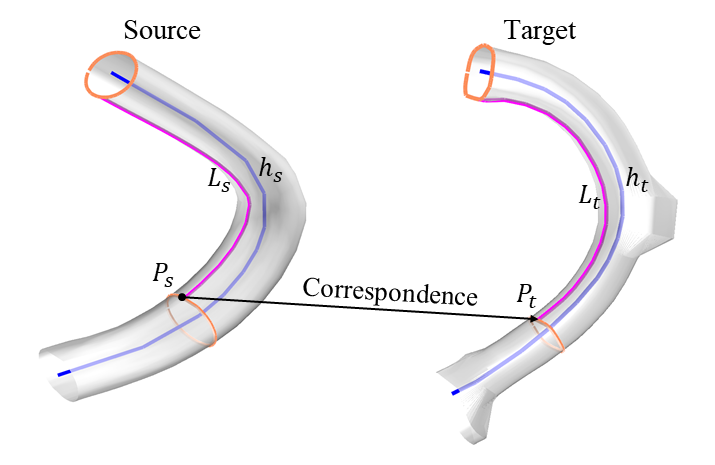}
  \caption{Illustration of our size-invariant correspondence to transfer a source point $p_s$ to a target point $p_t$ considering position consistency via surface distances (purple lines) and angle consistency via relative angles $\varphi$ on the corresponding cross sections (orange lines).
  }
  \label{fig:gc_method}
\end{figure}

As shown in Figure \ref{fig:teaser}, OakInk-based methods have issues with position and angle consistency. Specifically, the shape-interpolated correspondence fails to maintain the relative positions of finger contact regions as per the source pose, leading to the fingers either converging at a single point (Figure \ref{fig:teaser}a) or scattering erratically (Figure \ref{fig:teaser}b). On the other hand, the existing methods would distort the relative angles of the grasping fingers typically when transferring between two cross-category object parts.
For example, as illustrated in Figure \ref{fig:teaser}c, similar to the pot handle, the key is expected to be held between the thumb and forefinger on opposing sides. However, this transfer fails due to a significant semantic gap in finding the object correspondence.

In contrast, to preserve position and angle consistency, our method adopts two constraints to construct a GC-based correspondence between two object parts, i.e., keeping surface distance $L$ and relative angle unchanged. As described above, for a point $p$ on a GC surface, we first project it onto the GC skeleton from the opposite direction of the surface normal to acquire the project point $o$ and the skeleton height $h$, then represent the relative angle between $\overrightarrow{op}$ and the x-axis on $p$'s cross-section plane as $\varphi$. Thus, we transfer each source point $p_s(h,\varphi)$ on the source part to the target one $p_t$ following
\begin{equation}
    \varphi_{t} = \varphi_{s},
\end{equation}
\begin{equation}
    L_{t} = L_{s} + \Delta {H},
\end{equation}
where $\Delta {H}$ is an offset to adjust contact regions to be adaptive to source and target GCs. Empirically, we set
\begin{equation}
    \Delta {H} = (h_{t} - h_{s})/2.
\end{equation}
For surface distance $L$, we compute it depending on the surface along the shape skeleton given a surface point $p$, i.e.,
\begin{equation}
    L = \sigma (h, \varphi),
\end{equation}
where $\sigma (\cdot)$ denotes the 3D surface of an object part.

\subsection{Contact-based Hand Pose Synthesis}
\label{sec:pose_refine}

Given the transferred contact map on the target GC, we set the hand pose {that grasps the source part} as initialization and optimize it to approximate the contact points. We adopt an optimization method as in OakInk. We use contact consistency loss $E_\mathrm{consis}$ to maintain consistency between a hand and a contact map for grasping an object by attracting the anchors on the hand surface to its corresponding contact regions on the target part's contact map, and constraining the rotation axes and angles of the hands with the anatomical loss $E_\mathrm{anat}$. Instead of using the $\mathrm{SDF}$ of the shape that provides shape interpolations, we use the $\mathrm{SDF}$ of the whole target object to calculate hand-object interpenetration for a globally-based optimizing of hand and object.
\begin{equation}
    E_{\mathrm{intp}} = \sum_{V_{h,j}} -\min \left( \mathrm{SDF}_\mathrm{O}(V_{h,j}), 0 \right)
\end{equation}
The total loss is 
\begin{equation}
    L = E_{\mathrm{consis}} + E_{\mathrm{anat}} + E_{\mathrm{intp}}
\end{equation}

We use the Adam optimizer with 1000 iterations for refining a target hand pose.

\subsection{Computational Complexity Analysis}
The computational complexity of our \textit{GC-based Contact Transfer} technique depends on the number of source points $N$, the number of height values in the target $L$, and the complexity of surface height computation. Each point transfer involves a linear search through the target height values, resulting in a total complexity of $O(N \times L)$. Empirically, we set $N = 5000$, which is sufficient to obtain a representative contact map for hand poses.

Our experiment shows that \textit{GC-based Contact Transfer} takes 5-7 seconds and \textit{Contact-based Hand Pose Synthesis} takes 20-25 seconds for everyday objects in our experiment on a PC with Intel i9-9900K CPU and NVIDIA GeForce RTX 2080 Ti GPU. Besides, the \textit{Contact-based Hand Pose Synthesis} requires preprocessing with DeepSDF and takes around 40 minutes to train a DeepSDF model for an everyday object on the same PC. Thus, our approach is more suited for offline processing rather than real-time scenarios. Future work would focus on developing a contact-based hand pose synthesis method with less time requirement and implementing a faster version of \textit{GC-based Contact Transfer} with parallel programming to reduce the overall computational cost of our approach for potential interactive applications.

\section{Evaluation}
\label{sec:exp}

We have conducted extensive experiments to evaluate \sysName~quantitatively and qualitatively. We first introduce the dataset and metrics we chose to test our method and baselines in Section \ref{sec:dataset} and Section \ref{sec:metrics}.
We establish three hand pose transfer baselines, i.e., vanilla OakInk based on the global shape of objects, part-based OakInk (P-OakInk) that directly applies OakInk to object parts, and direct shape-based hand pose transfer (P-reg) for object parts. For ``P-reg'', we begin by applying shape-based fast global registration \cite{zhou2016fast} to estimate an initial transformation matrix between source and target object parts. We then refine this alignment using Iterative Closest Point (ICP) \cite{Rusinkiewicz2001Iterative}, which iteratively minimizes the distance between corresponding points to register the contact map from the source to the target object. Finally, we apply a similar optimization approach as our method to fine-tune the hand pose based on the refined contact map.
We show the comparisons between our method and the three baselines respectively for intra-category transfer (Section \ref{sec:intra_result}) and cross-category transfer (Section \ref{sec:cross_result}).

\subsection{Dataset}
\label{sec:dataset}
To evaluate our method and compare it with the three baselines, we selected 25 categories of 3D models from OakInk and 3 additional categories from online resources to prepare a dataset for HOI transfer. The dataset consists of source HOI data (including source objects and their corresponding grasping poses) and target objects. We randomly grouped source and target data that have similar affordances for the two objects, resulting in a total of 705 paired data for evaluation. Specifically, we set up 260 pairs for intra-category transfer (i.e., source and target objects from the same category) and 345 pairs for cross-category transfer.

\subsection{Metrics}
\label{sec:metrics}
Inspired by prior works \cite{yang_cpf:_2021, jiang_hand-object_2021, hasson_learning_2019, yang_oakink:_2022}, we utilize four metrics to evaluate the synthesized HOI data, including penetration depth, penetration volume, disjointed distance, and simulation displacement.

{\textit{Penetration Depth (PD).}} PD measures the depth that a
hand penetrates inside the surface of an object. We compute PD values as the maximum of the Euclidean distances between each penetrated hand vertex and its closest points on the object's surface.
 
{\textit{Penetration Volume (PV).}}
PV quantifies the volume of intersection between a hand pose and the grasping object. To calculate PV, we first voxelize the object shape using a voxel grid size of 128 via voxelization methods \cite{binvox,nooruddin03} and then count the number of hand-object intersected voxels. 

{\textit{Contact Ratio (CR).}} 
CR represents the ratio of the contact region where a hand grasps an object. Generally, a larger contact ratio indicates a more stable grasp as introduced in GraspTTA. We calculate CR by counting the ratio of contact points over the sampled points (5000 points) of a whole object.

{\textit{Disjointed Distance (DD).}} 
DD serves as an indicator of the stability of a hand grasp. We measure it by computing the average distance from the vertices of the five fingertips to their nearest points on the grasping object's surface.

{\textit{Simulation Displacement (SD).}} 
SD evaluates the physical stability of a grasp by simulating grasping objects using a physics simulation engine with gravity.
SD is measured as the slipping displacement of the object over a fixed duration during the simulation.

\begin{figure*}[h]
  \centering 
  \includegraphics[width=\textwidth, alt={multi}]{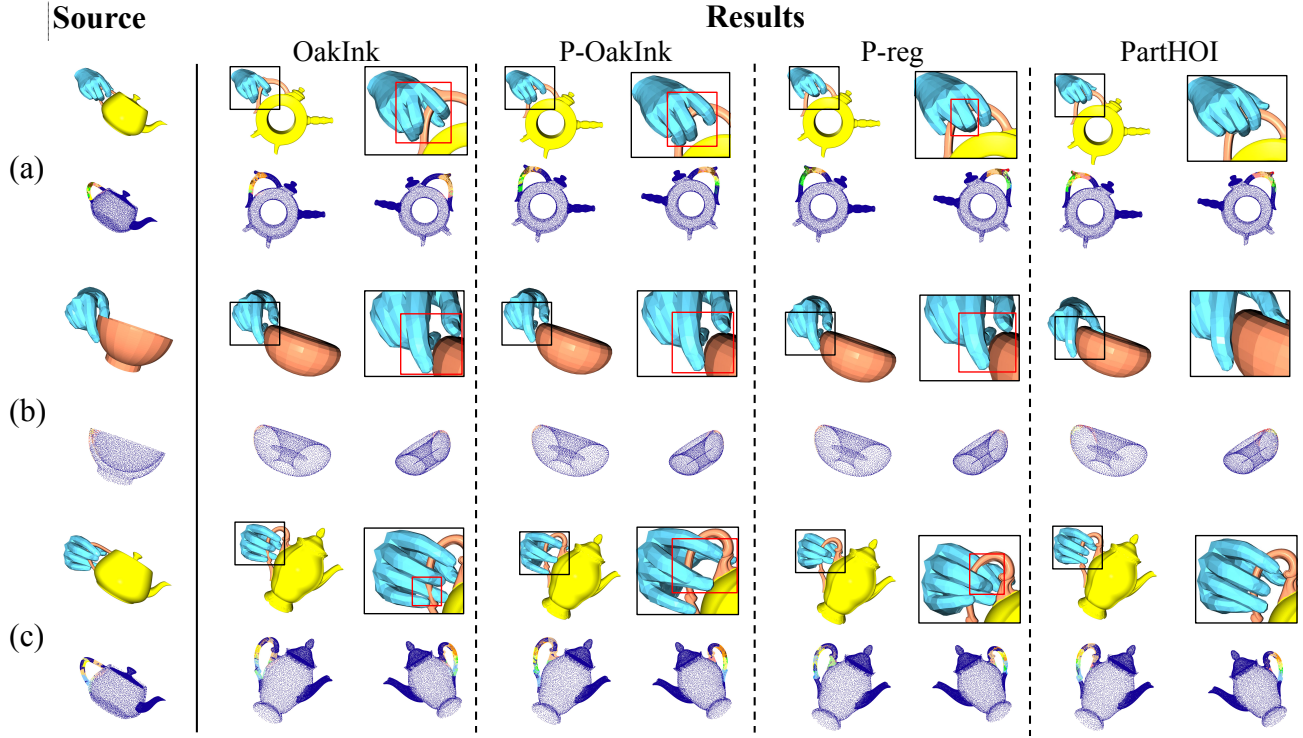}
  \caption{Qualitative results of our method compared to the three baseline methods for intra-category transfer.
  }
  \label{fig:intra}
\end{figure*}

\subsection{Intra-category Transfer}

\label{sec:intra_result}
\begin{table}[htb]
\center
\caption{Quantitative comparison of intra-transfer results.}
\begin{tabular}{lcccc}
\toprule 
Metrics
& Ours & OakInk & P-OakInk & P-reg
\\
\midrule
PD $cm$ $\downarrow$  & 0.183     & \bf{0.173} & 0.189 & 0.184\\
PV $cm^3$ $\downarrow$   & 1.423      & 1.191 & \bf{1.177} & 1.365\\
DD $cm$ $\downarrow$   & \bf{0.963}      & 1.080 & 1.075 & 1.025\\
CR $\%$ $\uparrow$ & \bf{24.94} & 24.34 & 24.68 & 24.26\\
SD 
$cm$ $\downarrow$   & \bf{2.581}      & 2.613 & 2.712 & 2.717\\
\bottomrule
\end{tabular}
\label{tab1}
\end{table}

Table \ref{tab1} shows the quantitative results of our method and the three baselines for intra-category transfer.
We test the methods on the intra-category subset and compute the average values of each metric across all the testing samples. It shows the hand poses produced by our method outperform those by the three baselines in terms of grasping stability (CR, DD, and SD). However, our method falls short regarding penetration measurement (PD and PV). This is because most of the intra-category objects have highly similar sizes and geometry, especially for local parts. Thus, the OakInk-based methods can smoothly interpolate and construct correspondence between the two shapes. Such correspondence can provide more dense and precise shape information to avoid hand-object penetration in intra-category cases.

Figure \ref{fig:intra} shows the qualitative results of intra-category transfer. Compared to the baselines, our method can generate more natural and stable hand poses.

\begin{figure*}[h!]
  \centering 
  \includegraphics[width=\textwidth, alt={multi}]{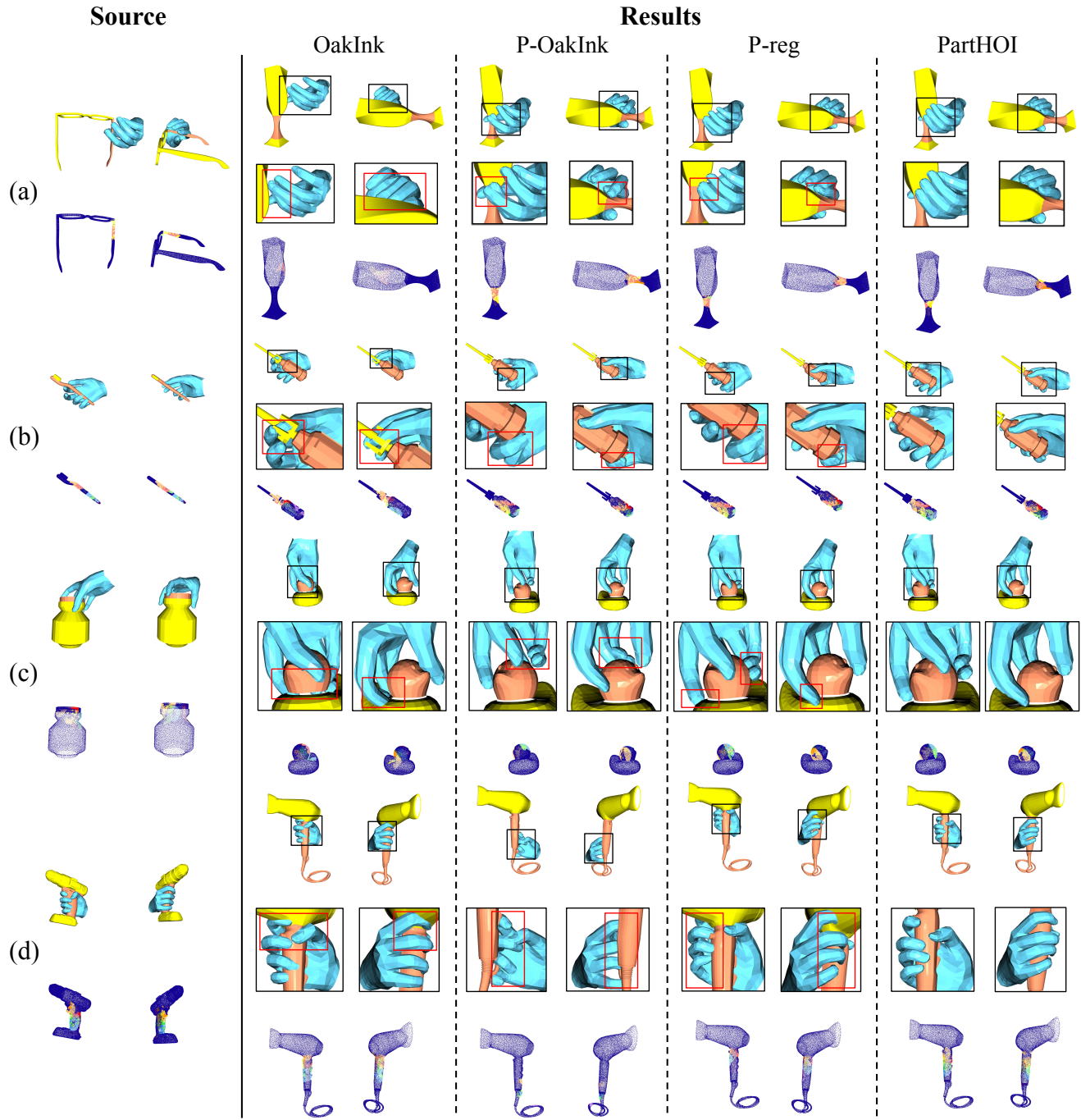}
  \caption{Qualitative results of our method compared to the three baselines for cross-category transfer of some objects.}
  \label{fig:cross}
\end{figure*}

\begin{figure*}[h!]
  \centering 
  \includegraphics[width=\textwidth, alt={multi}]{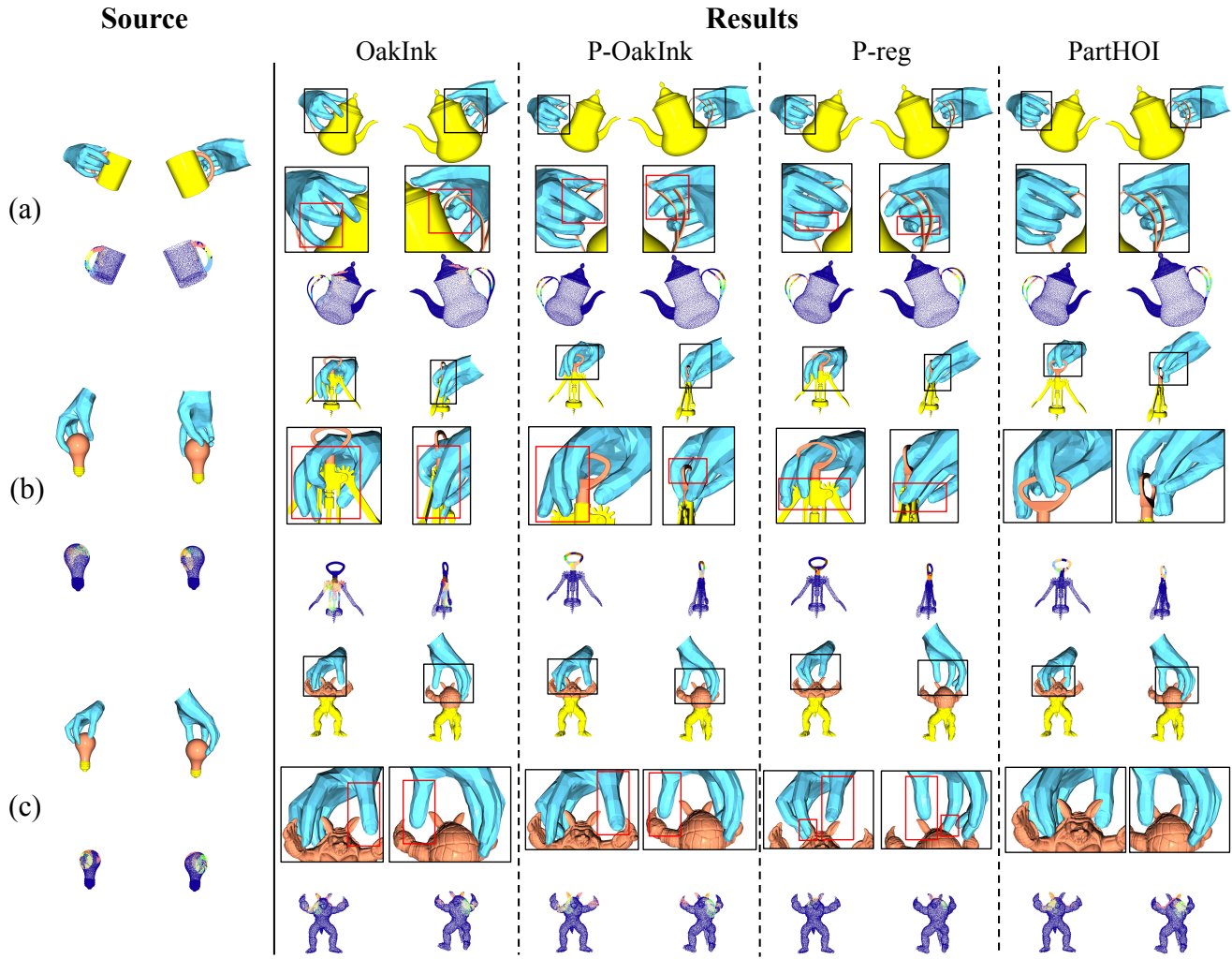}
  \caption{Qualitative results of our method compared to the three baselines for cross-category transfer of objects that require geometric abstraction.
  }
  \label{fig:cross_2}
\end{figure*}

\subsection{Cross-category Transfer}
\label{sec:cross_result}

\begin{table}[htb]
\center
\caption{Quantitative comparison of cross-transfer results.
}
\begin{tabular}{lcccc}
\toprule
{Metrics} & Ours & OakInk & P-OakInk & P-reg\\
\midrule
PD $cm$ $\downarrow$  & \bf{0.195}      & 0.334 & 0.197 & 0.312 \\
PV $cm^3$ $\downarrow$   & 1.619    & 4.158 & \bf{1.461} & 3.740\\
DD $cm$ $\downarrow$   & \bf{1.315}      & 3.796 & 1.530 & 1.752\\
CR $\%$ $\uparrow$ & \bf{28.80} & 21.79 & 27.96 & 27.89\\
SD $cm$ $\downarrow$   & \bf{3.051}      & 3.421 & 3.193 & 3.250\\
\bottomrule
\end{tabular}
\label{tab2}
\end{table}

Table \ref{tab2} shows the comparison for cross-category transfer. Compared to the three baselines, hand poses acquired by our method achieve the best quality in four metrics. The PD evaluation indicates that our method maintains high physical plausibility {by preventing significant hand penetration into the object's surface} for target hand poses. The DD, CR, and SD values reveal that our transferred hand poses have higher stability when grasping the target objects. 
However, our method exhibits lower PV performance compared to P-OakInk, which we attribute to our approach of abstracting shapes into generalized cylinders. This abstraction may wash away the minor structures on shape surfaces. As a result, transferring contact maps between two GCs would lack constraints from the details, leading to minor additional penetration. The results show that our method works significantly better than the baselines in cross-category transfer, especially compared to the performance in intra-category transfer. This is due to \sysName's part-based strategy reducing the influence of irrelevant object parts, and the GC-based correspondence preserving position and angle consistency across diverse categories.

Figures \ref{fig:cross} and \ref{fig:cross_2} show the qualitative results for cross-category transfer.
It shows that our method performs more robustly than the baselines for cross-category objects with large differences in shapes and sizes. Figure \ref{fig:cross_2} demonstrates that, even when some parts require geometric abstraction for approximation by generalized cylinders, our method can still generate high-fidelity hand poses.

\subsection{Perceptive User Study} To further evaluate the performance of our method, we conducted a perceptive user study to evaluate the visual naturalness and grasp stability.

We randomly selected HOI data respectively from the intra-transfer and cross-transfer datasets in Section \ref{sec:dataset}, with a total of 26=11+15 groups of the data. We generated the results by our method and the three baselines given the sampled HOI data.

The evaluation was conducted through an online questionnaire with a total of 35 participants, consisting of 26 questions. For each question, each participant was shown a source object and four pairs of generated results (a hand grasping the object) by the four methods with two multi-view rendered images. The results were displayed in random order. The participants were asked to rate each result on a Likert scale of 1 to 5 for visual naturalness and grasping stability (the higher, the better). In total, we obtained 35 (participants) $\times$ 26 (questions) = 910 subjective evaluations.

\begin{figure}[htb]
  \centering 
  \includegraphics[width=\columnwidth, alt={multi}]{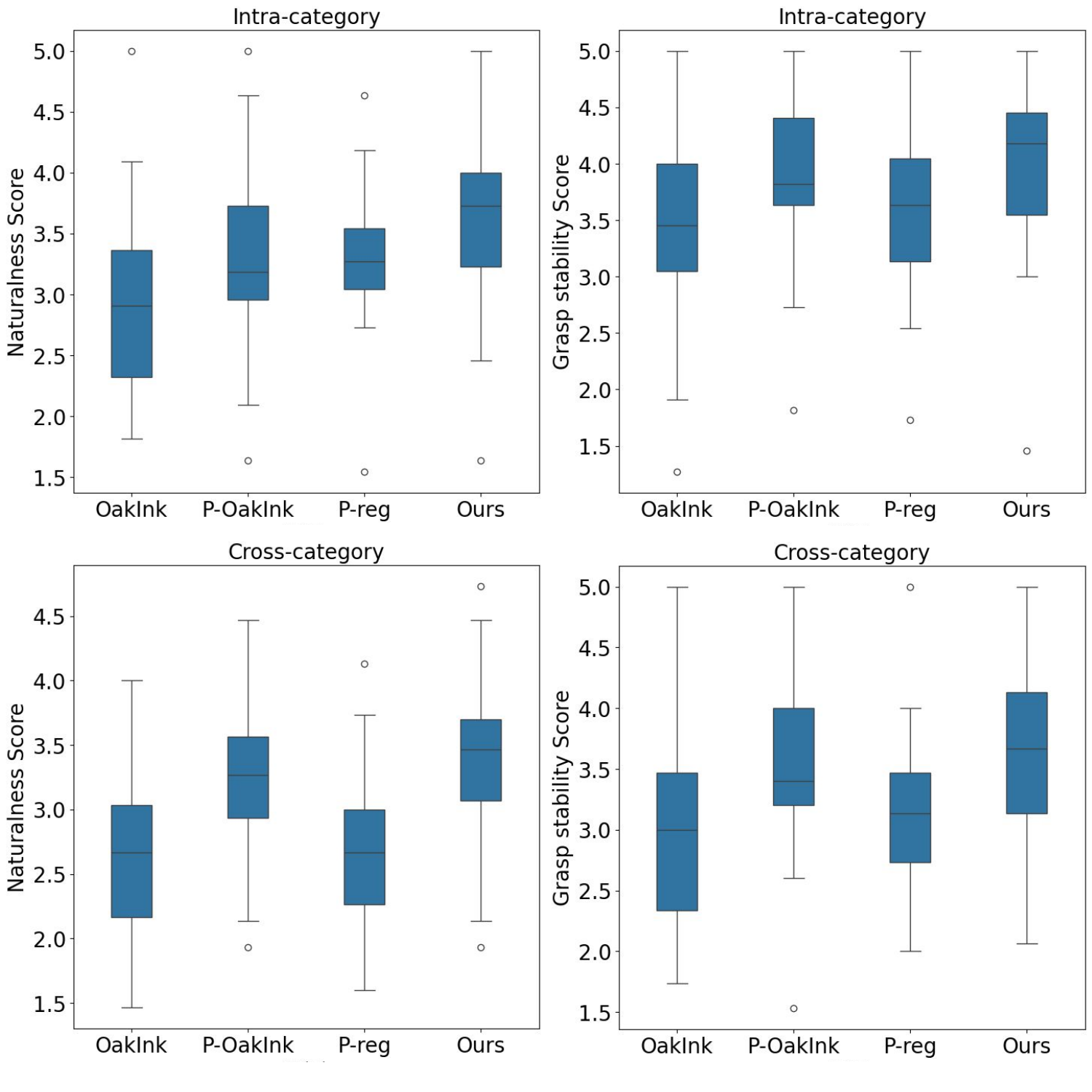}
  \caption{
Box plots of the average ratings of the naturalness and stability perceptive user study of the generated results for each method.
  }
  \label{fig:user_result}
\end{figure}

Figure \ref{fig:user_result} shows the statistics of the users' ratings. We conducted ANOVA tests (with $p < 0.001$) on the naturalness and grasp stability, and found significant effects for naturalness ($F = 6.180$) for intra-category transfer, and naturalness ($F = 14.482$) and grasp stability ($F = 8.520$) for cross-category transfer. The further paired T-tests (with $p < 0.001$) show that for intra-category transfer, our method (naturalness mean: 3.595) got significantly higher scores in the naturalness term than OakInk (naturalness mean: 2.906, $t = 6.190$) and P-OakInk (naturalness mean: 3.304, $t = 4.351$). For cross-category transfer, our method (naturalness mean: 3.396, stability mean: 3.647) got significantly higher scores in the naturalness term than all the three baseline methods and in the stability term than OakInk and P-reg. Paired T-tests for the three baseline methods compared to our method are OakInk (naturalness mean: 2.665, $t = 8.187$; stability mean: 2.973, $t = 7.117$), P-OakInk (naturalness mean: 3.217, $t = 3.803$), and P-reg (naturalness mean: 2.669, $t = 8.586$, stability mean: 3.149, $t = 6.729$).

\subsection{Multiple Parts Transfer}
\begin{figure}[htb]
  \centering 
  \includegraphics[width=\columnwidth, alt={multi}]{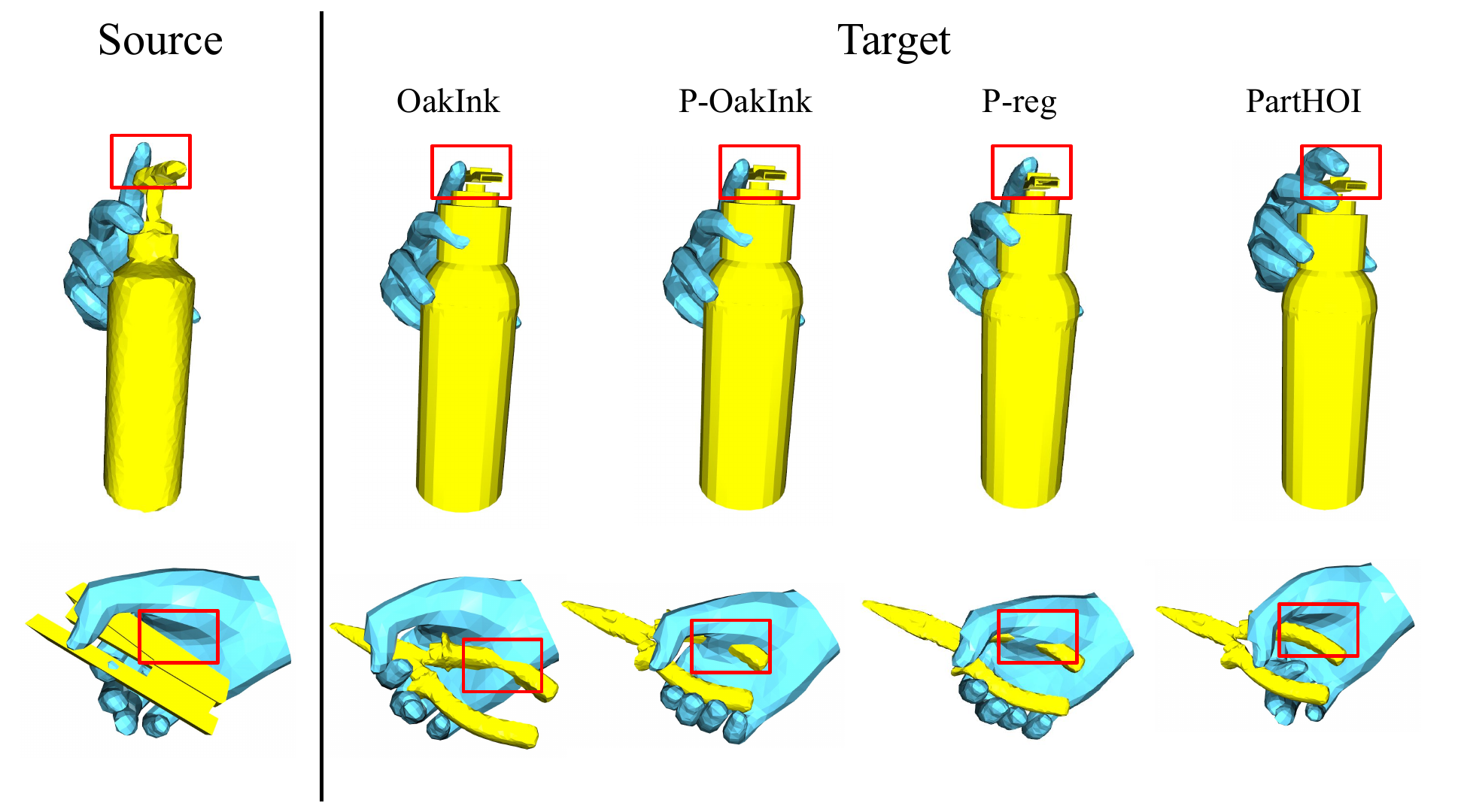}
  \caption{
 Comparison of \sysName~and the existing methods for transferring hand poses involving multiple parts.
  }
  \label{fig:multi_parts}
\end{figure}

For specific categories of objects, especially articulated objects (e.g., sprayers and pincers in Figure \ref{fig:multi_parts}), the contact affordance may involve multiple object parts. Directly applying OakInk or TOCH for such cases could not produce plausible results when there are significant differences in size, geometry, or object part layout between the source and target objects. For example, as shown in Figure \ref{fig:multi_parts}, OakInk-based methods or the correspondence-based method struggle to maintain contact with the sprayer's trigger when there are significant changes in their shape. Additionally, while the OakInk-based methods and the correspondence-based method can preserve plausible contact when transferring from a stapler to a pincer, they still result in unnatural poses and unintended penetration. In contrast, our method adopts unified GC-based correspondence between two object parts. This allows it to effectively handle multi-part grasp synthesis by transferring a contact map separately for each paired part. Figure \ref{fig:multi_parts} shows the hand poses generated by \sysName~can achieve the highest plausibility and the least penetration compared to the existing methods.

\section{Conclusions and Discussion}

\begin{figure}[htb]
  \centering 
  \includegraphics[width=\columnwidth, alt={multi}]{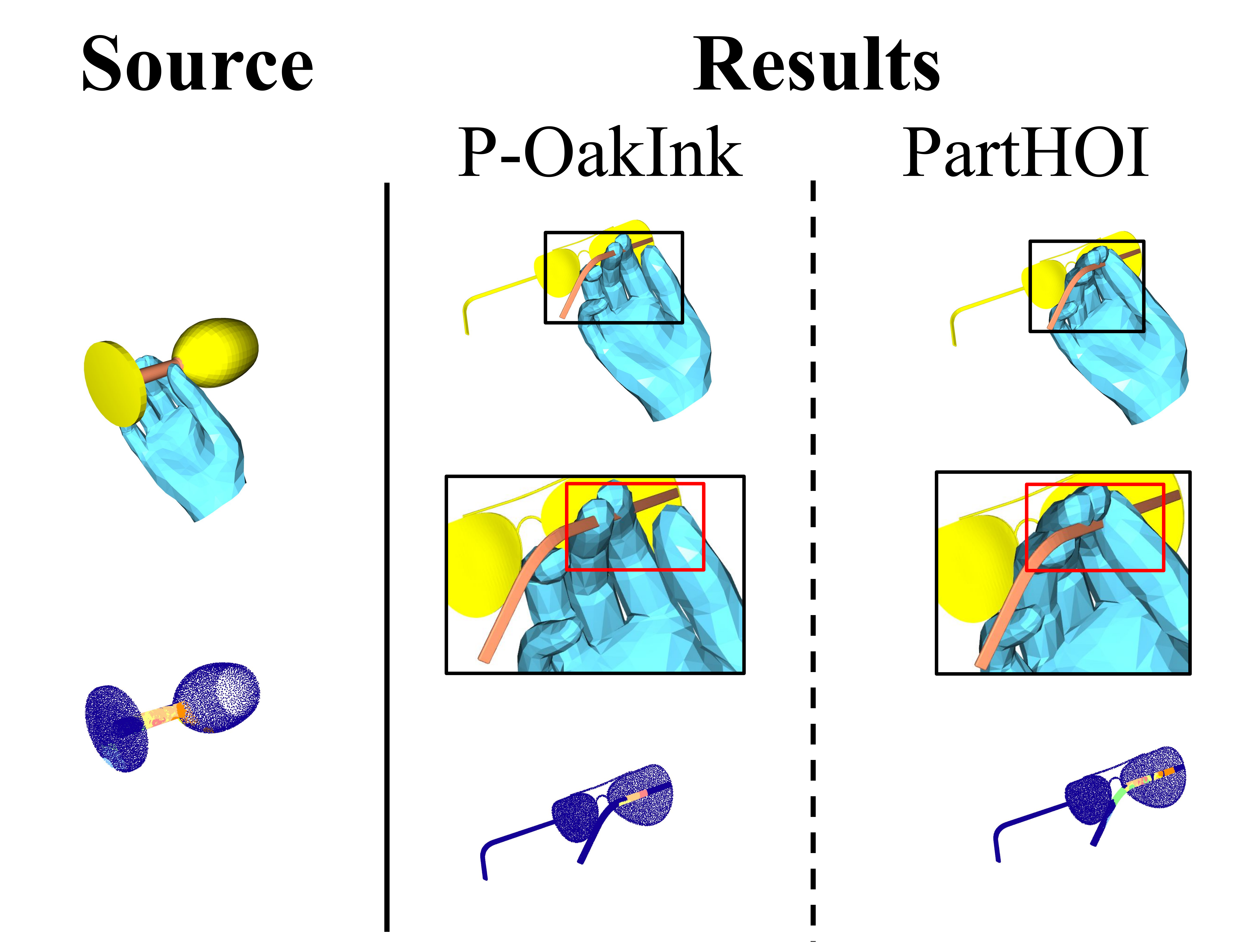}
  \caption{
 Failure cases of our method.
  }
  \label{fig:failure}
\end{figure}

In this work, we present \sysName, a part-based HOI transfer method, to produce plausible hand poses to grasp new objects given source HOI data. Based on the constructed GC-based correspondence, our method can transfer contact maps between two object parts with high shape and size differences. The extensive evaluation demonstrates that our method performs well in HOI transfer and is superior to the existing methods, especially for cross-category transfer. It implies that our method can effectively expand the scale of existing HOI datasets by propagating the existing hand poses to novel categories of objects. While our method works well for HOI transfer, there are some limitations. First, we manually segment grasping parts and extract their skeletons for GC modeling since existing automatic methods \cite{rosa_sig09, Huang2013, mo2019partnet} could not produce satisfactory results under diverse grasping intentions. In the future, we are interested in adapting such methods to our task. Second, our method may struggle to handle objects with highly complicated structures that GC could not model well. The abstraction of object parts tends to overlook minor structures, which can lead to inaccuracies when hands interact with these finer details. For example, our method does not handle scenarios where fingers interact with small-scale object features, such as inserting into tiny holes or pressing small buttons on surfaces. In the future, we can incorporate geometric descriptors (e.g., SIFT/SURF features) that capture local surface variations with a GC representation to build a more comprehensive correspondence. Third, as shown in Figure \ref{fig:failure}, our method can produce results with interpenetration when dealing with object parts that have very thin structures, such as the eyeglasses' frames. This issue arises because the signed distance values for these surfaces are too small to effectively prevent penetration during the optimization process. In future work, we plan to address this limitation by assigning greater weights to the signed distances of thin surfaces based on the GC parameters, thereby reducing interpenetration in the optimization step.

\subsection*{Declaration of availability of data and materials}

The code and datasets generated and analyzed during the current study will be available at https://github.com/waqc/PartHOI/tree/main.

\subsection*{Declaration of competing interest}

The authors have no competing interests to declare that are relevant to the content of this article.

\subsection*{Declaration of Funding}
This work was partially supported by the Centre for Applied Computing and Interactive Media (ACIM) of the School of Creative Media, CityU.

\subsection*{Declaration of authors' contributions}
Q. Wang: Conceptualization, Methodology, Software, Validation, Formal Analysis, Data Curation, Writing - Original Draft, Visualization\\
C. Xiao: Conceptualization, Methodology, Validation, Formal Analysis, Data Curation, Writing - Original Draft, Visualization, Supervision\\
M. Lau: Methodology, Formal Analysis, Writing - Review \& Editing, Supervision, Project Administration\\
H. Fu: Conceptualization, Methodology, Validation, Formal Analysis, Investigation, Resources, Writing - Review \& Editing, Supervision, Project Administration, Funding Acquisition

\subsection*{Acknowledgements}
We thank the anonymous reviewers for their constructive comments.

\bibliographystyle{CVMbib}
\bibliography{refs}

\end{document}